\pdfoutput=1

\documentclass[11pt]{article}

\usepackage{authblk}

\usepackage{EACL2023}

\usepackage{times}
\usepackage{latexsym}
\usepackage{bm} 

\usepackage[T1]{fontenc}

\usepackage[utf8]{inputenc}

\usepackage{microtype}

\usepackage{graphicx}
\usepackage{latexsym}
\usepackage{algorithm, amsmath, amsfonts, amsthm, mleftright, multirow, booktabs}
\usepackage{siunitx}
\hypersetup{breaklinks=true}

\usepackage{caption}

\theoremstyle{definition}

\usepackage{algpseudocode}

\usepackage{microtype}

\usepackage{comment}

\title{Self-Adaptive Named Entity Recognition by Retrieving \\ 
Unstructured Knowledge}

\author[$\dag\ddag$]{\textbf{Kosuke Nishida}}
\author[$\S$]{\textbf{Naoki Yoshinaga}}
\author[$\dag$]{\textbf{Kyosuke Nishida}}
\affil[$\dag$]{NTT Human Informatics Laboratories, NTT Corporation}
\affil[$\ddag$]{The University of Tokyo}
\affil[$\S$]{Institute of Industrial Science,
        the University of Tokyo}
\affil[$\dag$]{\texttt{\{kosuke.nishida.ap, kyosuke.nishida.rx\}@hco.ntt.co.jp}}
\affil[$\S$]{\texttt{ynaga@iis.u-tokyo.ac.jp}}

\date{}

\newcommand{\yn}[1]{\textcolor{black}{#1}}
\newcommand{\ner}{\textsc{ner}}
\newcommand{\nerbert}{\textsc{nerbert}}
\newcommand{\ukb}{\textsc{ukb}}
\newcommand{\realm}{\textsc{realm}}
\newcommand{\saner}{\textsc{sa-ner}}
\newcommand{\conll}{\textsc{c}o\textsc{nll}03}

\begin{document}
\maketitle
\begin{abstract}
Although named entity recognition ({\ner}) helps us to extract domain-specific entities from text (\textit{e.g.}, artists in the music domain), it is costly to create a large amount of training data or a structured knowledge base to perform accurate {\ner} in the target domain. Here, we propose self-adaptive {\ner}, which retrieves external knowledge from unstructured text to learn the usages of entities that have not been learned well. To retrieve useful knowledge for {\ner}, we design an effective two-stage model that retrieves unstructured knowledge using uncertain entities as queries. Our model predicts the entities in the input and then finds those of which the prediction is not confident. Then, it retrieves knowledge by using these uncertain entities as queries and concatenates the retrieved text to the original input to revise the prediction. Experiments on CrossNER datasets demonstrated that our model outperforms 
strong baselines by 2.35 points in F$_1$ metric.
\end{abstract}

\section{Introduction}

Named entity recognition ({\ner}) helps us to extract entities from text
\yn{in various domains} such as 
biomedicine~\cite{ner_gene}, disease~\cite{ner_disease}, and COVID-19~\cite{ner_covid19}. 
However, 
accurate neural
{\ner} requires a massive amount of training data~\cite{ner_data3, ner_data1, ner_data4}.
As well, the annotation of a domain-specific {\ner} dataset costs a lot of money because it requires the involvement of domain experts.

To 
compensate for the lack of training data in {\ner}, \yn{researchers have utilized} external knowledge.
Traditional feature-based {\ner} 
uses \yn{features based on} gazetteers or name lists~\cite{gazetteer4, gazetteer3, gazetteers2} as external knowledge.
Although recent neural
{\ner} methods can even benefit from 
gazetteers and name lists~\cite{gazetteer_nn3, gazetteer_nn1, gazetteer_nn2}, only a few domains with structured knowledge bases (gazetteers) have this merit. 
Thus, several studies have resorted to
using raw text (unstructured knowledge) \yn{to perform}
weakly-supervised learning 
on general-domain structured knowledge~\cite{ner_weak, gazetteer_nn2, ner_bert}.


\begin{figure}
\centering
		\includegraphics[width=\linewidth]{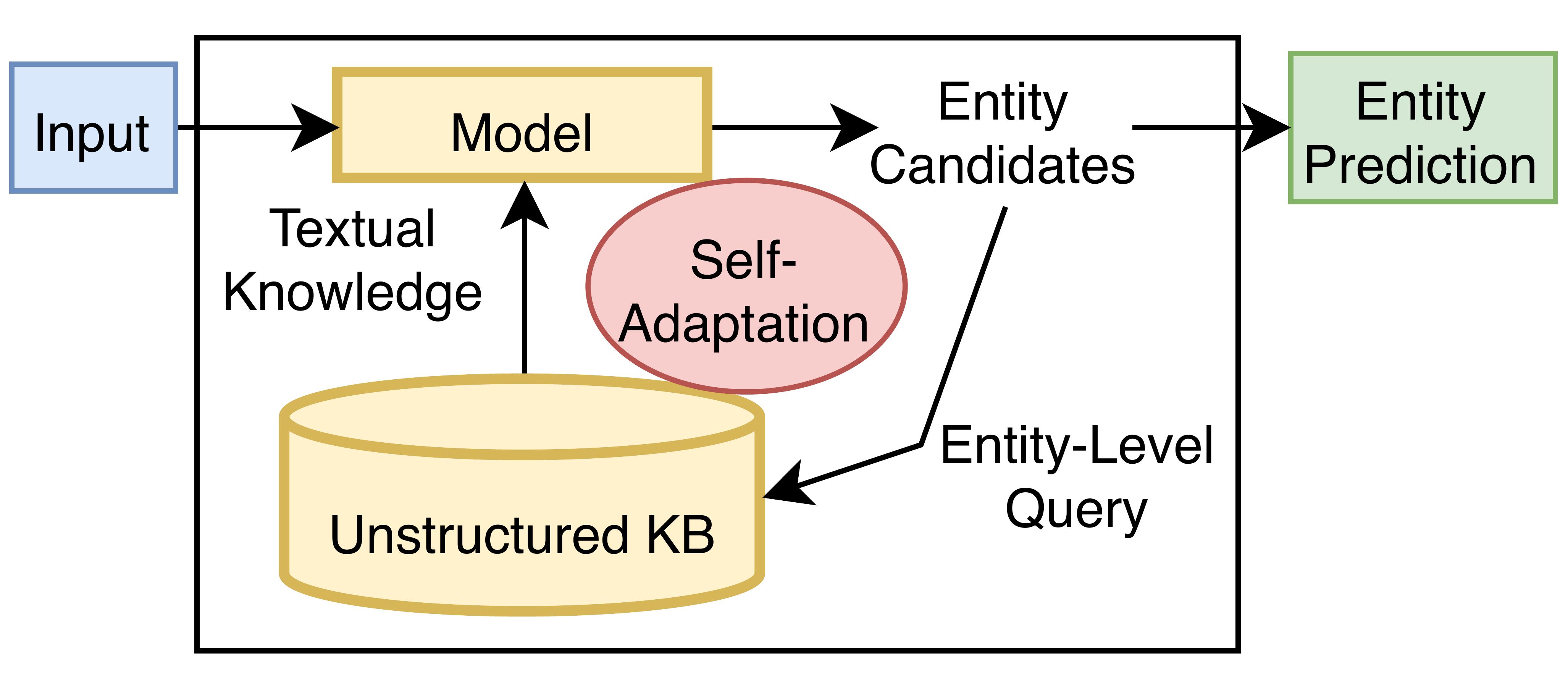}
		\caption{Concept of self-adaptive {\ner}: \yn{the model predicts entity candidates to conduct entity-level retrieval from the unstructured KB\@; then it revises the prediction with reference to the retrieved knowledge.}}
		\label{fig:top}
\end{figure}


In this paper, we explore the potential of utilizing unstructured knowledge in the {\ner} task by referring to it
at inference time. 
Our basic 
idea is inspired by recent retrieval-augmented language models (\textsc{lm}s)~\cite{realm}.
These models are pre-trained with retrieval-augmented masked language model (\textsc{mlm}), so that they can perform well in open-domain question answering (\textsc{odqa}) by retrieving relevant unstructured knowledge using a question as a query.
However, as we will later confirm in the experiments, the models designed for \textsc{odqa} are not effective in the {\ner} task because it requires an understanding of many entities in the input text.

To deal with this problem, we propose a retrieval-augmented model capable of determining which entities to focus on in the input text for knowledge retrieval. 
The proposed \textbf{self-adaptive {\ner} ({\saner}) with unstructured knowledge model} searches an unstructured knowledge base ({\ukb}) when it lacks confidence in its prediction. 
We create the {\ukb} automatically by splitting a raw text corpus into pieces and assigning dense vectors as keys to each piece of unstructured knowledge.
To help in understanding local semantics, we design a retrieval system tailored for {\ner}; our model predicts the entities and then retrieves knowledge in terms of those it is not confident in predicting. 

To evaluate our method's capability of retrieving 
\yn{useful}
knowledge about entities, we conducted experiments on various 
{\ner} datasets~\cite{conll03, finance_ner, crossner}, some of which 
\yn{have domain-specific 
types.}

Our contributions are summarized as follows:
\begin{itemize}
    \item We are the first to integrate retrieval-augmentation into {\ner}. {\saner} retrieves entity-level knowledge dynamically for {\ner}. 
    \item In experiments, {\saner} outperformed strong baselines pre-trained in a supervised and self-supervised fashion by 1.22 to 2.35 points.
    \item We reveal why knowledge retrieval is useful for {\ner}.  
    We found that our model is effective
    on entities not included in 
    the general-domain pre-training dataset.
\end{itemize}

\section{Task Settings}
We developed {\saner} 
\yn{to solve the problems of {\ner} 
with unstructured knowledge.}
{\ner} is a sequence tagging task in which the model inputs a token sequence $X \in V^L$, where $V$ is the vocabulary and $L$ is the maximum sequence length. 
The model outputs a BIO label sequence of the same length. Let $C$ be the number of types. Then, the number of the BIO labels is $2C+1$.

{\saner} assumes a corpus as an unstructured knowledge, which is split into token sequences of length $L$, following the existing retrieval-augmented language model (\textsc{lm})~\cite{retro}, in order to store a large corpus efficiently. We retrieve $m$ pieces of knowledge and concatenate them into $X$.
We feed the concatenated text $X^+ \in V^{(m+1)L}$ to the model.

\section{Related Work}
\yn{Here, we review
{\ner} that uses raw text (unstructured knowledge)
without structured knowledge, with in-domain structured knowledge, with general-domain structured knowledge, and for pre-training of billion-scale \textsc{lm}s .}
Also, we review the retrieval-augmented \textsc{lm}s.

\subsection{{\ner} with unstructured knowledge}

\yn{Researchers have utilized various clues to retrieve useful raw text for {\ner}. Traditional {\ner} models focus on surrounding contexts~\cite{ner_global1, ner_global2, ner_global3} and linked documents~\cite{twitter_link} to capture non-local dependencies.}
More recent neural {\ner} models benefit from
neighbor sentences 
\yn{to obtain better contextualized word representations~\cite{ner_corpus2,ner_corpus1}. Meanwhile,
 \citet{knowledgener} and \citet{li-etal-2020-unified} encode knowledge contexts on entity types such as questions, definitions, and examples taken from in-domain structured \textsc{kb}s (\textit{e.g.,} \textsc{umls} Meta-thesaurus).
In this study, we developed a generic method that retrieves useful raw text (unstructured text) for \textsc{ner}.}

\yn{Distant supervision~\cite{distant} uses} structured knowledge to annotate raw text with pseudo labels. Performing distantly supervised fine-tuning with in-domain structured knowledge after the \textsc{mlm} pre-training is effective in domain-specific {\ner}~\cite{distant_chem,distant_bio}.
However, domain-specific distant supervised learning \yn{depends on} 
the structured knowledge's coverage of the label set
of the downstream task. 

Weakly supervised learning with general-domain structured knowledge~\cite{ner_weak, bond, gazetteer_nn2, ner_bert} can transfer general-domain knowledge to the target domain. Its methods learn the entity knowledge through weakly supervised learning, even though the target task has domain-specific entities and types~\cite{ner_bert}.
We confirmed that our model achieved a performance gain by using raw text as unstructured knowledge at inference time because the world knowledge cannot be stored in the limited-sized model.

\textcolor{black}{Pre-trained \textsc{lm}s memorize  factual knowledge in their models through pre-training on unstructured corpus~\cite{lm_kb1, lm_kb2, lm_kb3}. Recently, billion-scale generative pre-trained \textsc{lm}s have been proposed~\cite{t5, gpt-3}. 
Although the generative models cannot be applied naively to structured prediction tasks such as {\ner}, some papers tackled {\ner} with the generative \textsc{lm}s~\cite{ner_gen1, ner_gen2, ner_gen3, ner_gen4}. One of the advantages of retrieval-augmented \textsc{lm}s over billion-scale \textsc{lm}s is ease of maintenance; For instance, the models can use up-to-date Wikipedia as the \textsc{ukb}s.}

\subsection{Retrieval-Augmented Language Models}
\textsc{lm}s using external knowledge have recently been proposed~\cite{realm, rag, fid, emdr, retro}.
However, they focus on language modeling and \textsc{odqa}, and successful retrieval-augmented \textsc{lm}s in {\ner} have not been reported.
They obtain queries for knowledge retrieval in such a way that each query represents the whole input or a fixed-length chunk split from the input.
Therefore, they cannot retrieve knowledge that tells 
the usages of the entities, which is important for {\ner}.
In addition, because an input may include many entities, 
the model should focus on only those entities whose knowledge is not stored in the model. However, retrieval-augmented \textsc{lm}s have not incorporated such a mechanism to create and filter multiple queries.

\citet{retreive_train1} and \citet{retreive_train2} found that retrieving knowledge from the training data is also useful, as it provides knowledge not stored in the trained model. Therefore, we implemented {\saner} in such a way that it uses both labeled and unlabeled {\ukb}s.

\begin{figure*}
\centering
		\includegraphics[width=0.96\textwidth]{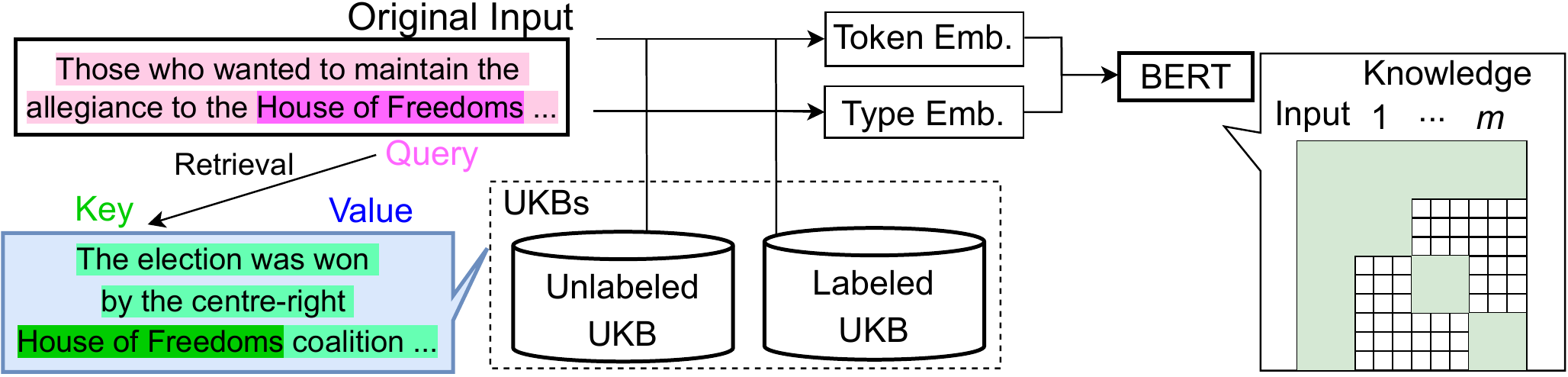}
		\caption{\yn{Overview of our self-adaptive {\ner} with 
		knowledge retrieval from {\ukb}s, which store text with n-gram and sentence embeddings as keys.
		The labeled {\ukb} has 
		text with labels encoded as token type embeddings.} The queries are embeddings of unconfident entities and input. 
		We use a sparse matrix in the self-attention modules in \textsc{bert}.}
		\label{fig:model}
\end{figure*}

\yn{\citet{tome}} used a virtual knowledge base 
\yn{whose values are vector representations.
Focusing on entity knowledge, they extracted mentions from hyperlinks in Wikipedia to learn their representations. They reported that the virtual \textsc{kb} was less accurate but more
efficient than \textsc{fid}~\cite{fid}, 
which reads the input and textual knowledge with attention.}

\section{Method}
Here, we present {\saner}\@.
We explain the construction of the unstructured knowledge base (\S\ref{ssec:kb}), the encoder architecture (\S\ref{ssec:encoder}), the two-stage \textsc{ner} algorithm which
revises the prediction using the unstructured knowledge  (\S\ref{ssec:inf}), the training method (\S\ref{ssec:training}), and the pre-training method (\S\ref{ssec:pre}).

\subsection{Unstructured KB Construction}
\label{ssec:kb}
We create an unlabeled {\ukb} from raw text and a labeled {\ukb} from the training data.
We assume in-domain \yn{text} 
as a source of unlabeled unstructured knowledge and split it into token sequences of length $L$, which is equal to the maximum length of the {\saner} inputs. 
In addition, following \citet{retreive_train1}, we add the model's training data as labeled unstructured knowledge. We set $L=64$ to avoid truncating most of the original inputs.


The unstructured knowledge is stored in the {\ukb}s with associated keys.
The keys of the sequence are the sentence embedding and the n-gram embeddings.
\citet{bert_sentemb} showed that the average of the token embeddings is more useful for sentence embedding than the first \texttt{[CLS]} embedding and that the embeddings in the lower layers are also important, as well as those in the last layer.
Therefore, we define the sentence embedding and n-gram embedding as the average pooling of the token representations. 
The token representations are the concatenations of the frozen \textsc{bert} input and output, so that both the context-free and contextualized meanings are considered.

To \yn{select only} entity-like n-grams as the keys, 
we remove those n-grams that have stop words or have no capital letters.
In addition, we use string matching for filtering.
We hold only the knowledge that includes the n-grams appearing in the training data for the {\ukb}s used at training time. Also, we hold the knowledge that includes the n-grams appearing in the training or development (test) data for the {\ukb}s at the inference on the development (test) data.
\yn{Instead of string matching, we can use a summarization-based filtering for n-gram keys, as detailed in Appendix~\ref{append:summarize}.
We formulate the extraction of a fixed number of representative n-grams from a sequence as an extractive summarization.
We use a sub-modular function as the objective~\cite{summarization_submodular}; thus, the greedy algorithm has a $(1-1/e)$ approximation guarantee.}

Following \citet{retreive_train1}, we use the labeled {\ukb} even in 
training to reduce the training-test discrepancy; in such case, the model does not retrieve the 
input itself from the labeled {\ukb}.

\subsection{Encoder}
\label{ssec:encoder}
We use \textsc{bert}~\cite{bert} and a linear classifier with a softmax activation as the encoder $f$.
Figure~\ref{fig:model} shows the encoder structure.
To represent the label information from the labeled knowledge base in the model, we provide additional token-type embeddings.
Though the token type is always zero in the conventional \textsc{bert} model for {\ner}, we use $2C+3$ token-type IDs;
\[
t_i =
\begin{cases}
0 & \text{if $x_i$ is the original input} \\
1 & \text{if $x_i^+$ is unlabeled knowledge} \\
l_i + 2 & \text{if $x_i^+$ is labeled knowledge} \\
\end{cases},
\]
where $l_i$ is the label of the labeled knowledge, and $X^+$ is the concatenated text.

In the self-attention module, we use the sparse attention technique to reduce the space and time complexity from $\mathcal{O}(m^2L^2)$ to $\mathcal{O}(mL^2)$.
As shown in Figure~\ref{fig:model}, we mask the inter-knowledge interaction.\footnote{
We implement $X^+$ as an $(m+1) \times L$ tensor. We calculate three attention matrices: intra-sequence attention ($(m+1) \times L \times L$), knowledge-to-input attention ($m \times L \times L$), and input-to-knowledge attention ($m \times L \times L$). This operation takes advantage of parallel computing on the GPU.
}
Let $k$ be a function that returns 0 as the sentence id if the sequence is the input $X$ and $1,...,m$ if the sequence is the knowledge.
Accordingly, the attention matrix before the softmax operation is 
\[
A_{ij} =
\begin{cases}
\frac{\bm{Q}_i^\top \bm{K}_j}{\sqrt{d_k}} & \text{if $k(i) =k(j)$ or $k(i)k(j) =0$} \\
- \infty & \text{otherwise}
\end{cases},
\]
where $i,j$ are the token indices, $d_k$ is the number of dimensions of the attention head, and $\bm{Q}$ and $\bm{K} \in \mathbb{R}^{(m+1)L \times d_k}$ are query and key matrixes.


\subsection{Two-stage Tagging of Self-Adaptive {\ner}}
\label{ssec:inf}
{\saner} performs two-stage tagging, \textit{i.e.,} calculation of $\bm{P}=f(X)$, and calculation of $\bm{P}^+=f(X^+)$.
The purpose of the first stage is to 
find the entities that require additional information and 
obtain queries for knowledge retrieval.
The second stage is to \yn{refine} 
the labels with the retrieved knowledge.
The motivation behind this design is to retrieve useful entity-wise knowledge to disambiguate individual tokens in {\ner}.
We predict the entity spans for entity-level retrieval.
We use only the unconfident entities as the \yn{entity-based} queries in order to exclude
\yn{unnecessary knowledge from the retrieved results.}
The pseudo-code of the model is listed in Algorithm~\ref{alg}.

We obtain the classification probabilities of the given text $\bm{P} =f(X) \in \mathbb{R}^{L \times (2C+1)}$ or that of the text with knowledge $\bm{P}^+ =f(X^+) \in \mathbb{R}^{L \times (2C+1)}$, where the vectors after position $L$ are ignored.
The model parameters are shared in the two stages.

\paragraph{First Stage}
We collect unconfident entities $\mathcal{U}$ in $X$ and feed $X$ to the model to obtain the classification probability $\bm{P} \in \mathbb{R}^{L \times (2C+1)}$.
Then, we extract the entities $\mathcal{E}$ from $X$ in accordance with the predicted labels $\hat{\bm{y}} =\mathrm{argmax}_c \bm{P}_{\cdot c} \in  \mathbb{N}^{L}$. 
The confidence score of a predicted entity $e$ is $c_e =\min_{i\in I_e} P_{i, \hat{y}_i}$, where $I_e$ is 
\yn{the span of 
$e \in \mathcal{E}$.}
If the type predictions are inconsistent in an entity (\textit{e.g.}, [B-LOC, I-PER]), we set $c_e =0$.
We collect the unconfident entities $\mathcal{U} \subseteq \mathcal{E}$ whose confidence scores
are less than a 
threshold $\lambda_{\mathrm{conf}}$.


\begin{algorithm}[t!]
\small
	\caption{\yn{Two-stage self-adaptive {\ner}}}\label{alg} 
	\begin{algorithmic}[1]
		\Require input $X$, \textsc{kb}s, hyperparamters $m, \lambda_{conf}$
		\State Predict probability $\bm{P}=f(X)$
		\State Compute confidence score $c_e =\min_{i\in I_e} P_{i, \hat{y}_i}$ for each predicted entity $e \in\mathcal{E}$ with span $I_e$
		\State Obtain unconfident entities $\mathcal{U} =\{e | e \in \mathcal{E}, c_e < \lambda_{\mathrm{conf}} \}$
		\State Add the sentence and unconfident-entity embeddings to the queries, $Q$
		\State Initialize the retrieval results $R =\Phi$
		\For {query $q_i$ in the queries, $Q$}
		\State Retrieve $m$ nearest-neighbor keys for $q_i$ from the \textsc{kb}s
		\State Store their values with the distance in $R$
		\EndFor
		\State Deduplicate $R$ to obtain top-$m$ knowledge $K_1^m$ from $R$
		\State Output probabilities $\bm{P}$ and $\bm{P}^+=f(X^+ =[X; K_1^m])$
	\end{algorithmic}
\end{algorithm}

Then, we obtain the queries, which are the sentence 
and entity embeddings.
The sentence embedding is the average pooling over all token embeddings. 
Each unconfident entity $u \in \mathcal{U}$ has multiple entity embeddings: average-pooled vectors of n-grams which share at least one token with $u$.
The n-grams are filtered out similarly as in the {\ukb} construction (\S~\ref{ssec:kb}). 
$E$ denotes the number of 
\textit{entity embeddings} (which are embeddings of n-grams overlapping with $u \in \mathcal{U}$).
Each token embedding is a concatenation of the \textsc{bert} input and output.
Note that we only consider sentence-to-sentence and entity-to-n-gram matching.
We retrieve the top-$m$ nearest neighbors of the sentence embedding from the sentence embeddings in the {\ukb}s and of the entity embeddings from the n-gram embeddings.
Then, we select the top-$m$ nearest knowledge from the collected $2(E+1)m$ knowledge while 
\yn{deduplicating} the backbone knowledge sequence by keeping the knowledge having the minimum distance.

\paragraph{Second Stage}
We concatenate the knowledge $K_1^m$ to the input $X$ and obtain the classification probability $\bm{P}^+ =f(X^+ = [X; K_1^m])$.
Finally, the model outputs \textsc{bio} labels in accordance with
$\bm{P}$ 
for the tokens in the confident entities and in accordance with $\bm{P}^+$ for the other tokens.

\subsection{Training}
\label{ssec:training}
To train our two-stage {\saner}, we utilize supervision on the training data to refine unconfident entities and design the loss function.

\paragraph{Unconfident Entity Collection}
In the training phase, we add the misclassified entities, i.e., those of which the prediction is not correct, to the unconfident entities $\mathcal{U}$ described in \S\ref{ssec:inf}.

\paragraph{Loss Function}
\yn{We use two cross-entropy losses, $\mathcal{L}_{1}$ for the model prediction without knowledge (the first step) and $\mathcal{L}_{2}$ for the model prediction with knowledge  (the second step).
The total loss function is $\mathcal{L}_{2} + \lambda_{1} \mathcal{L}_{1}$, where $\lambda_{1}$ is a hyperparameter.}

\subsection{Pre-training} 
\label{ssec:pre}
As is done in retrieval-augmented language models for \textsc{odqa}~\cite{realm, retro}, we add a retrieval-augmented pre-training stage before the fine-tuning. We propose two methods for NER-aware retrieval-augmented pre-training. The first method uses a general domain NER dataset, CoNLL03~\cite{conll03}.
The model is pre-trained with the method described above~
(\S\ref{ssec:kb}\textasciitilde\S\ref{ssec:training}).

The second method involves a large-scale self-supervised pre-training following {\nerbert}~\cite{ner_bert}.
Although the {\ukb} in {\saner} and the pre-training data overlapped in some cases, {\saner} can use the knowledge effectively by referring to it at inference time.

\begin{table*}[t!]
\centering
    \small
    \tabcolsep3pt
		\begin{tabular}{l|cccccc|cc} \toprule
                & AI. & Mus. & Lit. & Sci. & Pol. & Avg. & Fin. &  CoNLL03 \\ \midrule
                \# Train (\# \textsc{ne} types) & 100 (14) & 100 (13) & 100 (12) & 200 (17) & 200 (9) & --- & 1169 (4) & 14987 (4) \\ 
                \textsc{bert}$^\dag$ & 50.37 & 66.59 &  59.95 & 63.73 & 66.56 &  61.44 & --- & --- \\
                \textsc{dapt}$^\dag$ & 56.36 & 73.39 & 64.96  & 67.59 & 70.45 & 66.55 & --- & --- \\
                {\nerbert}$^\ddag$ & 60.39 & 76.23 & 67.85 & 71.90 & 73.69  & 70.01 & --- &  --- \\ \midrule
                \textsc{bert} {\scriptsize on \conll} & 56.97 {\scriptsize (1.05)} & 69.10 {\scriptsize (1.08)} & 64.37 {\scriptsize (0.73)} & 65.76 {\scriptsize (0.58)} & 70.16 {\scriptsize (0.56)}  & 65.27 {\scriptsize (0.80)} & 72.35  {\scriptsize (5.32)} & ---  \\
                {\realm}-{\ner} {\scriptsize on \conll} & 58.05 {\scriptsize (1.15)} & 71.17 {\scriptsize (0.63)} & 64.58 {\scriptsize (0.69)} & 66.33 {\scriptsize (0.66)} & 69.38 {\scriptsize (0.36)} & 66.56 {\scriptsize (0.80)} & 70.03  {\scriptsize (1.35)} & --- \\
                {\saner} {\scriptsize on \conll} & \textbf{60.31} {\scriptsize (1.03)} & \textbf{72.20} {\scriptsize (0.79)}  & \textbf{66.23} {\scriptsize (1.30)} & \textbf{68.22} {\scriptsize (0.57)} & \textbf{71.18} {\scriptsize (0.57)}  & \textbf{67.62} {\scriptsize (0.85)} & \textbf{74.02}  {\scriptsize (2.29)} & ---  \\ \midrule
                \textsc{bert} {\scriptsize on {\nerbert}} & 62.05 {\scriptsize (0.66)} & 76.45 {\scriptsize (0.90)} & 69.68 {\scriptsize (0.26)} & 72.10 {\scriptsize (0.67)} & 74.38 {\scriptsize (0.40)} & 70.93 {\scriptsize (0.58)} & 75.05 {\scriptsize (7.47)} & 90.25 {\scriptsize (0.11)}  \\
                {\realm}-{\ner} {\scriptsize on {\nerbert}} & 64.32 {\scriptsize (0.31)} & 77.55 {\scriptsize (0.69)} & 70.42 {\scriptsize (0.60)} & 72.52 {\scriptsize (0.42)} & 74.45 {\scriptsize (0.38)} & 71.85 {\scriptsize (0.43)} & 73.34 {\scriptsize (1.74)}  & 89.94 {\scriptsize (0.42)} \\
                {\saner} {\scriptsize on {\nerbert}} & \textbf{65.27} {\scriptsize (0.95)} & \textbf{78.71} {\scriptsize (0.47)} & \textbf{71.79} {\scriptsize (0.57)} & \textbf{74.38} {\scriptsize (0.19)} & \textbf{74.63} {\scriptsize (0.36)} &   \textbf{72.96} {\scriptsize (0.51)} & \textbf{75.77} {\scriptsize (1.01)} & \textbf{90.49} {\scriptsize (0.49)}  \\ \bottomrule
                \end{tabular}
	\caption{Main results on the test set. The model was pre-trained on CoNLL03 or the {\nerbert} dataset from the BERT-base-cased model. We ran five experiments with different seeds. Standard deviations are parenthesized. Performances of the previous models are cited from~\citet{crossner}$^\dag$ and~\citet{ner_bert}.$^\ddag$
	\textsc{bert} on {\nerbert} corresponds to our implementation of the {\nerbert} model.}
	\label{tab:main_result}
\end{table*}

\begin{table*}[t!]
\centering
    \small
    \tabcolsep3pt
		\begin{tabular}{l|cccccc|cc} \toprule
                & AI. & Mus. & Lit. & Sci. & Pol. & Avg. & Fin. &  CoNLL03 \\ \midrule
                \textsc{d}istil\textsc{bert} {\scriptsize on \conll} & 54.16 {\scriptsize (1.21)} & 66.64 {\scriptsize (0.54)} & 60.53 {\scriptsize (1.26)} & 64.14 {\scriptsize (0.49)} & 67.61 {\scriptsize (0.70)} & 62.61 {\scriptsize (0.84)} & 68.78  {\scriptsize (6.35)} & ---  \\
                {\realm}-{\ner} {\scriptsize on \conll} & 53.85	{\scriptsize (1.38)} & 67.03 {\scriptsize (0.41)} & 61.83 {\scriptsize (1.38)} & 64.19	{\scriptsize (0.17)} & 69.09 {\scriptsize (0.52)} & 63.20 {\scriptsize (0.54)} & 70.35  {\scriptsize (5.04)} & --- \\
                {\saner} {\scriptsize on \conll} & \textbf{55.31} {\scriptsize (1.03)}  & \textbf{67.25} {\scriptsize (1.14)}  & \textbf{61.53} {\scriptsize (1.18)}  & \textbf{65.71} {\scriptsize (1.03)}  & \textbf{69.36} {\scriptsize (0.55)}  & \textbf{63.83} {\scriptsize (0.99)} & \textbf{72.89} {\scriptsize (2.71)} & ---  \\ \midrule
                \textsc{d}istil\textsc{bert} {\scriptsize on {\nerbert}}  & 59.52	{\scriptsize (0.89)} & 71.60 {\scriptsize (1.05)} & 63.52 {\scriptsize (0.47)} & 69.26 {\scriptsize (0.97)} & 68.88 {\scriptsize (0.64)} & 66.56 {\scriptsize (0.80)} & 73.36  {\scriptsize (4.17)} & 89.23  {\scriptsize (0.19)} \\
                {\realm}-{\ner} {\scriptsize on {\nerbert}} & 60.39 {\scriptsize (0.53)} & 71.39 {\scriptsize (0.33)} & 62.89 {\scriptsize (0.19)} & 68.18 {\scriptsize (0.83)} & 69.79 {\scriptsize (0.82)} & 66.53 {\scriptsize (0.54)} & 74.35 {\scriptsize (5.06)} & 88.54  {\scriptsize (0.62)} \\
                {\saner} {\scriptsize on {\nerbert}} & \textbf{61.90} {\scriptsize (0.38)} & \textbf{73.61} {\scriptsize (0.45)} & \textbf{65.48} {\scriptsize (0.31)} & \textbf{70.44} {\scriptsize (0.69)} & \textbf{69.95} {\scriptsize (0.90)} & \textbf{68.27} {\scriptsize (0.55))} & \textbf{75.40}  {\scriptsize (1.46)} & \textbf{89.50}  {\scriptsize (0.30)} \\ \bottomrule
                \end{tabular}
	\caption{Main results on the test set. The models were pre-trained from DistilBERT-base-cased.}
	\label{tab:main_result_distil}
\end{table*}

\paragraph{\nerbert}
The pre-training corpus is Wikipedia\@. If the consecutive words in the corpus have a hyperlink, the words are labeled as an entity. We categorize such entities with the DBpedia Ontology~\cite{dbpedia}. If the entity exists in the ontology, we categorize it to its type. If it does not exist or it belongs to multiple types, we categorize it to the special ``ENTITY'' type.

We split the corpus into fixed-length token sequences,\footnote{Although the original {\nerbert} uses a sentence as a unit, we use a fixed length in order to share the setting with {\ukb}s.} and 
\yn{extract the sequences with tokens labeled with the DBpedia types.}
We reduce the proportion of ``ENTITY'' labels by using filtering rules and down sampling.
The resulting dataset has 33M examples, 939M tokens, and 404 types.

We add a final linear layer with a trainable parameter $\bm{W}_{\mathrm{pre}} \in \mathbb{R}^{d \times (2C_{\mathrm{pre}} +1)}$ to the top of \textsc{bert}, where $d$ is the hidden size of \textsc{bert} and $C_{\mathrm{pre}}$ is the number of types. Before fine-tuning, the final 
layer is replaced with a randomly initialized linear layer whose output dimension is determined by the downstream task.
\yn{Refer to Appendix~\ref{append:data} and the original paper~\cite{ner_bert} for details.}

\paragraph{Knowledge Retrieval}
We use the {\saner} model in the pre-training to \yn{reduce the pre-training and fine-tuning discrepancy.}
We use the pre-training data itself as {\ukb}s. We retrieve knowledge with its pseudo-labels from the data as labeled knowledge and 
randomly delete the pseudo-labels to make the knowledge unlabeled. We set the deletion probability as $0.95$ to simulate downstream tasks where the unlabeled {\ukb} is larger than the labeled {\ukb}.
For efficiency, 
we use Wikipedia hyperlinks as the keys and queries of the retrieval. Instead of a two-stage prediction, we sample $m$ pieces of knowledge that includes an entity in the original input.

\section{Evaluation}
We conducted experiments on three {\ner} datasets to evaluate the effectiveness of our self-adaptive {\ner} with unstructured knowledge. We used the entity-level F$_1$ as the metric, following the literature.

\subsection{Dataset}

\smallskip\noindent\textbf{CrossNER}~\cite{crossner} consists of five domains: politics, science, music, literature, and AI\@. This small-scale dataset was created by annotating the sentences extracted from the Wikipedia articles in each domain. It provides the textual corpus extracted from Wikipedia for the in-domain pre-training. We used it for the unstructured {\ukb}.\footnote{We can see if the self-adaptive {\ner} is useful even though the unlabeled knowledge overlaps the {\nerbert} pre-training data. Also, we report the effect of overlapping entities in the pre-training data and CrossNER dataset on the performance in Appendix~\ref{append:overlap}}
The label sets are different among the domains. 

\smallskip\noindent\textbf{Finance}~\cite{finance_ner} 
is a medium-scale {\ner} dataset 
collected from U.S\@. 
SEC filings. We used the Wikipedia articles in the finance domain as the textual corpus $\mathcal{D}$ to construct the unlabeled \ukb\@. The label set is \yn{person, organization, location, and miscellaneous}.

\smallskip\noindent\textbf{CoNLL03}~\cite{conll03} is a widely used large-scale {\ner} dataset collected from Reuters news stories between August 1996 and August 1997. 
We used the Reuters-21578 text classification dataset~\cite{reuter21578}, which was collected from Reuters in 1987, as $\mathcal{D}$.
The label set is the same as that of Finance.

\subsection{Compared Models}
Our text encoder and tokenizer were the pre-trained \textsc{bert}-base-cased model~\cite{bert} or \textsc{d}istil\textsc{bert}-base-cased model~\cite{distilbert}. 
 All experiments used the hyperparameters determined on the development set of CrossNER-Politics; refer to Appendix~\ref{append:hyper}. 

We pre-trained the compared models on the {\conll} or {\nerbert}~\cite{ner_bert}\footnote{\yn{Our implementation was different from the original {\nerbert} in terms of the fixed length sequences, initialization, loss function, and data collection results; refer to Appendix~\ref{append:hyper}.}} datasets before fine-tuning. In addition to the \textsc{bert} model (\textit{i.e.}, \textsc{bert} with {\conll} or {\nerbert} pre-training), we implemented the {\ner} version of {\realm} ({\realm}-{\ner}).
For {\realm}-{\ner}, we replaced the retrieval-augmented \textsc{mlm} of {\realm} with 
our retrieval-augmented pre-training methods tailored for {\ner} to assess the effectiveness of our knowledge retrieval. Also, we set $m =1$, removed the entity-level retrieval, and ignored the labeled {\ukb}.
We cited the results of the previous models: \textsc{bert}, {\nerbert}, and \textsc{dapt}~\cite{tapt}, which is the domain-adapted \textsc{bert} baseline.\footnote{We did not cite the results of {\nerbert} on Finance because the authors did not report the data splits.} We compared our model with models consisting of \textsc{bert} and a linear classifier because the classifier architecture is out of the scope of our study.



\subsection{Main Results}

Table~\ref{tab:main_result} and Table~\ref{tab:main_result_distil} show the main results.
The proposed model outperformed the baselines across all target domains, models, and pre-training datasets.
The improvement is typically larger in the lower-resource domain with more types, because \yn{per-type supervision is limited in such case.}

\paragraph{\yn{Does self-adaptive {\ner} improve the performance of the \textsc{ner}-aware pre-training?}}
{\saner} outperformed \textsc{bert} with {\conll} and {\nerbert} pre-training. This indicated that the self-adaptation using unstructured knowledge at inference time has the effect of obtaining additional knowledge that is not stored in the model, even though the model has seen the unstructured knowledge in the pre-training.
Moreover, because we can increase the unlabeled {\ukb} after pre-training, the model can acquire new knowledge more efficiently than by conducting additional pre-training.

\paragraph{Does self-adaptive {\ner} improve the performance of the retrieval-augmented LM baseline?}
{\saner} outperformed {\realm}-{\ner}.
{\saner} retrieves knowledge with the entity-level retrieval from the labeled and unlabeled {\ukb} and encodes large pieces of knowledge due to the sparse attention. 
These techniques improved the usefulness of the knowledge for {\ner}\@. The contributions of each component are discussed in the ablation studies.
We also found that {\realm}-{\ner} tends to be not good in the setting \# Train $>1000$. Because {\realm}-{\ner} retrieves a piece of knowledge with only the sentence-level query, knowledge retrieval is not always useful in that setting.


\subsection{Ablation Studies}
\begin{table}[t!]
\centering
	\small
    \tabcolsep2.3pt
		\begin{tabular}{lcc} \toprule
		        Method & Acc & $\Delta$ \\ \midrule
                Proposed & 77.33 {\scriptsize (0.19)} & \\ \cmidrule(l{1pt}r{0pt}){1-3}
                \, w/o Entity-level Retrieval & 76.21 {\scriptsize (0.23)} & $1.12$  \\
                \, w/o Sentence-level Retrieval & 76.54 {\scriptsize (0.48)} & $0.79$ \\ 
                \cmidrule(l{1pt}r{0pt}){1-3}
                \, w/o Confident Entities (\textit{i.e.,} $\lambda_{\mathrm{conf}} >1$) & 76.91 {\scriptsize (0.24)} & $0.42$ \\
                \, w/o using First-Step Prediction on $\mathcal{E} \setminus \mathcal{U}$ & 76.97 {\scriptsize (0.33)} & $0.36$ \\
                \cmidrule(l{1pt}r{0pt}){1-3}
                \, w/o Unlabeled Knowledge & 76.23 {\scriptsize (0.44)} & $1.10$ \\
                \, w/o Labeled Knowledge & 76.82 {\scriptsize (0.45)} & $0.51$\\
                \midrule
                {\nerbert} & 75.90 {\scriptsize (0.22)} & 1.43 \\ 
                \bottomrule
                \end{tabular}
	\caption{Ablation studies on the development set of the politics domain. $\Delta$ shows the drop from the proposed model. 
	Each ablation was conducted in the fine-tuning and evaluation.}
	\label{tab:ablation}

\end{table}
Table~\ref{tab:ablation} shows the results of the 
ablation studies. We used the best performing {\saner} with {\nerbert} pre-training as the full model.
We found that all components of {\saner} improved performance.

\paragraph{Does the entity-level retrieval improve performance?}
First, we confirmed the usefulness of self-adaptive knowledge retrieval, because knowledge retrieval based on the model's entity prediction is more useful for {\ner} than conventional sentence-level retrieval ($\Delta 1.12$ vs. $\Delta 0.79$). 
Also, we found that both knowledge retrievals improve {\ner} performance.

\paragraph{Does the distinction about confidence improve the performance?}
Second, we investigated the efficacy of distinguishing the predicted entities in terms of confidence.
The model retrieves knowledge about unconfident entities $\mathcal{U} =\{e | c_e < \lambda_{\mathrm{conf}}, e \in \mathcal{E} \}$, 
and then refines the prediction for only the unconfident entities with the retrieved knowledge.
We set $\lambda_{\mathrm{conf}} >1$ to remove the distinction.
We observed that ignoring confident entities in creating queries is slightly effective ($\Delta 0.42$), because we can restrict the retrieval results to informative knowledge for {\ner}\@. 
Then, we used the second-step prediction for all tokens. We found that reusing the first-step prediction for confident entities 
improved performance slightly ($\Delta 0.36$). 
Using the first-step prediction is important for confident entities because the retrieved knowledge is \yn{likely to be} irrelevant to them. 
We consider that making the distinction is more useful in the smaller $m$ setting where the amount of knowledge is limited.

\paragraph{Do the labeled and unlabeled {\ukb}s improve the performance?}
Finally, we confirmed that both the labeled and unlabeled {\ukb}s are important ($\Delta 1.10$ and $\Delta 0.51$). 
The unlabeled {\ukb} covers various contexts, and the labeled {\ukb} has supervision. The \yn{two types of} {\ukb} have different roles in helping the model recognize entities.


\subsection{Discussion}

\paragraph{Does the performance of our model depend on the amount of knowledge?}
\begin{figure}
\centering
		\includegraphics[width=\linewidth]{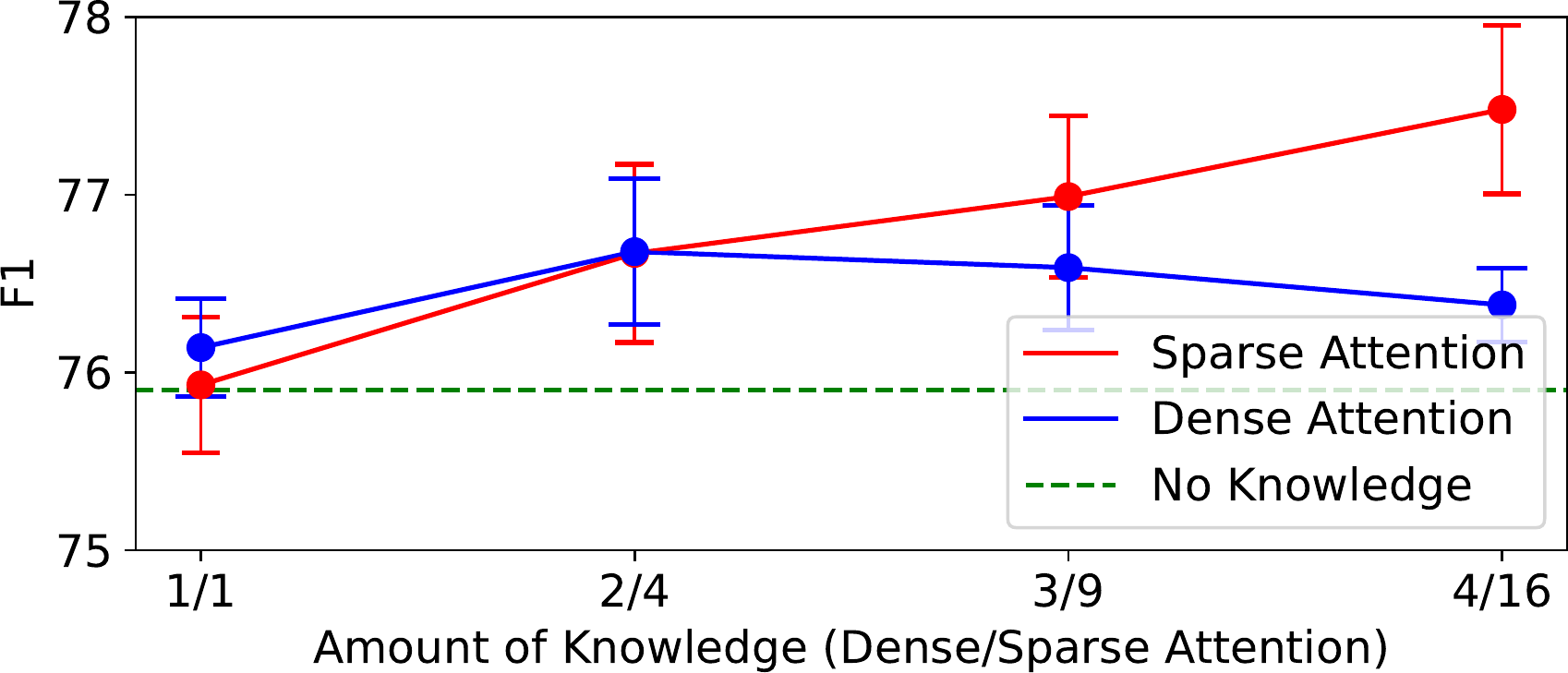}
		\caption{\yn{F$_1$ score versus the amount of knowledge $m$. The error bars show the standard deviation over five runs. 
		For a fair comparison of the time and space complexity, we compared the methods with different values of $m$, 
		since our sparse attention runs in $\mathcal{O}(ml^2)$ while 
		the dense attention runs in $\mathcal{O}(m^2l^2)$}.}
		\label{fig:num}
\end{figure}
Figure~\ref{fig:num} plots F$_1$ score versus the amount of knowledge $m$.
We can see that more pieces of knowledge led to higher F$_1$ scores. Because the time and space complexity of the sparse attention is linear in the number of pieces of knowledge, the sparse attention is suitable for large $m$. However, the dense attention did not improve performance in the case of large $m$. We consider that the sparse attention represents the intra- and inter-sequence interactions more effectively than the naive dense attention can.

\begin{table}[t!]
\centering
    \small
    \tabcolsep3pt
		\begin{tabular}{lrcc} \toprule
                & \# Entities & {\nerbert} & Proposed \\ \midrule
                All & 3472 & 75.90 {\scriptsize (0.22)} & 77.33 {\scriptsize (0.19)} \\
                Seen in Training & 661 & 84.05 {\scriptsize (1.43)} & 85.20 {\scriptsize (0.21)} \\
                Unseen in Training & 2811 & 71.39 {\scriptsize (0.29)} & 73.03 {\scriptsize (0.36)} \\
                Seen in Pre-Training & 3083 & 77.58 {\scriptsize (0.17)} & 78.83 {\scriptsize (0.29)}\\
                Unseen in Pre-Training & 389 & 50.90 {\scriptsize (1.63)} & 54.18 {\scriptsize (1.85)} \\ \bottomrule
                \end{tabular}
	\caption{Detailed results on the development set.}
	\label{tab:discussion}

\end{table}

\paragraph{What types of entity require external knowledge?}
Table~\ref{tab:discussion} lists the results for when the target entities were restricted to each type, which is defined in terms of whether the supervision of an entity was included in the training and pre-training data.
The proposed model outperformed {\nerbert} on all types.
The improvement was 1.15 points for the ``seen in training'' type and 1.64 points for the ``unseen in training'' type.
Therefore, self-adaptation has an effect regardless of whether or not the entity exists in the training data; we also observed this effect in the ablation studies.



Regarding the ``unseen in pre-training'' type, the proposed model improved performance by 3.28 points.
The pre-training dataset collected from Wikipedia shares a lot of entities in the CrossNER dataset created from Wikipedia, and thus whether the tokens are labeled as entities in the pre-training dataset (i.e., the tokens have Wikipedia hyperlinks) has a large effect on performance.
We confirmed that PRE-training data is more valuable than one might think, similarly to the findings of \citet{retreive_train1} that the reference to the training data at inference time is worthwhile.

\paragraph{\textcolor{black}{Is the self-adaptive {\ner} sensitive to the unconfidence threshold?}}

\textcolor{black}{
To investigate the sensitivity of \textsc{sa-ner} to the hyperparameter, we set $\lambda_{\mathrm{conf}}$ to various values at inference time after we trained the model with $\lambda_{\mathrm{conf}} =0.9$.} 

\textcolor{black}{
Table~\ref{tab:robustness} shows the results. The performance is on par if  $\lambda_{\mathrm{conf}} \in [0.8, 0.95]$. Therefore, \textsc{sa-ner} is not sensitive to $\lambda_{\mathrm{conf}}$. We also confirmed that modifying the prediction of the high-confidence entities is harmful ($\lambda_{\mathrm{conf}} =1$) and thus using $\lambda_{\mathrm{conf}}$ is useful. Moreover, we observed that modifying the prediction of certain entities (3.6\% of the total number) is important. These entities are ones in which the token-level predictions were inconsistent, and their confidence $c_e$ were set to 0.
}

\begin{table}[t]
\centering
\small
\begin{tabular}{llr}
\toprule
$\lambda_{\mathrm{conf}}$ & Acc & {\scriptsize Unconfident Proportion} \\ \midrule
0     & 76.21 {\scriptsize (0.23)} & 0.00\%   \\
0.1   & 77.14 {\scriptsize (0.30)}  & 3.60\%   \\
0.5   & 77.18 {\scriptsize (0.23)}  & 5.38\%   \\
0.7   & 77.21  {\scriptsize (0.28)} & 9.60\%   \\
0.8   & 77.30   {\scriptsize (0.21)} & 12.25\%  \\
0.9   & \textbf{77.33} {\scriptsize (0.19)} & 16.63\%  \\
0.95  & \textbf{77.33} {\scriptsize (0.16)}  & 21.04\%  \\
0.97  & 77.24  {\scriptsize (0.19)} & 24.74\%  \\
0.99  & 77.13 {\scriptsize (0.16)}  & 31.99\%  \\
0.995 & 77.12  {\scriptsize (0.17)} & 37.61\%  \\
0.999 & 77.01  {\scriptsize (0.22)} & 63.63\%  \\
1     & 76.91  {\scriptsize (0.24)} & 100.00\% \\
\bottomrule
\end{tabular}
\caption{\textcolor{black}{F1 score versus confidence threshold $\lambda_{\mathrm{coef}}$. Unconfident proportion indicates the proportion of unconfident entities to all entities. We omitted rows $0.2, 0.3, 0.4,$ and $0.6$, whose performance is the same as that of the rows directly above.}}
	\label{tab:robustness}
\end{table}

\paragraph{Does the self-adaptive {\ner} depend on the filtering 
method
of the n-grams?}
\begin{table}[t!]
\centering
    \small
		\begin{tabular}{lc} \toprule
                String Matching Filtering & 77.33 {\scriptsize (0.19)} \\ 
                w/o Knowledge Retrieval ({\nerbert}) & 75.90 {\scriptsize (0.22)} \\
                w/o Entity-level Retrieval & 76.21 {\scriptsize (0.23)} \\ \cmidrule(lr){1-2}
                Summarization-based Filtering  & 77.02 {\scriptsize (0.40)} \\
                \bottomrule
                \end{tabular}
	\caption{Performance of self-adaptive {\ner} with summarization-based filtering of n-gram embeddings.}
	\label{tab:sum}

\end{table}

We compared the two filtering methods for n-gram embeddings in the {\ukb}.
The string matching method used the information of the n-grams appearing in the training or development (test) splits in the evaluation on the development (test) set.
The summarization-based method 
\yn{just set}
the maximum number of n-grams in each piece of knowledge.

Table~\ref{tab:sum} shows the results. Both methods outperformed the no-knowledge baseline (\nerbert) and the ablated model without the entity-level knowledge retrieval. 
The summarization-based filtering requires fewer assumptions and is  computationally efficient, \yn{although it is less accurate}.

\subsection{Qualitative Analysis}
\begin{table}[t!]
\centering
\tabcolsep2.2pt
    \small
		\begin{tabular}{lp{170pt}}\toprule
		Input & 
		the Association for the Rose in the Fist of Lanfranco
		  Turci and those who wanted to maintain the
		  allegiance to the \textcolor{blue}{House of Freedoms} coalition. \\ 
		  \cmidrule(l{0.2em}r{0.2em}){1-2}
		Knowledge & 
		The election was won in Sardinia by the centre-right \textcolor{blue}{House of Freedoms} coalition ... voted party with 30.2\% .\\ 
		\cmidrule(l{0.2em}r{0.2em}){1-2}
		Prediction & organization $\rightarrow$ political party \\ \midrule
		Input & 
		Director Michael Moore partnered with producers Harvey Weinstein and \textcolor{blue}{Bob Weinstein} in May 2017 to produce and distribute Fahrenheit 11/9 . \\ 
		\cmidrule(l{0.2em}r{0.2em}){1-2}
		Knowledge & 
		... \textcolor{blue}{Bob Weinstein}, the founders of Miramax Films. \\ 
		\cmidrule(l{0.2em}r{0.2em}){1-2}
		Prediction & politician $\rightarrow$ politician \\ \bottomrule
                \end{tabular}
	\caption{Qualitative Analysis. 
	\yn{One representative piece of knowledge retrieved for the input is provided.}}
	\label{tab:qualitative}

\end{table}

Table~\ref{tab:qualitative} shows examples of our model. The first example is a case in which the self-adaptation improved the model prediction.
The original input itself does not have evidence that the House of Freedoms is a political party. However, the knowledge provides this evidence by mentioning it in the context of an election.
The second example is the most common fault in the political domain. Because of the imbalance between the training labels of person and politician, the person entities tend to be misclassified as politician entities. Although both the input and the knowledge indicate that Bob Weinstein is not a politician, the model made the wrong prediction.

\section{Conclusions}
We proposed {\saner}, which is designed for {\ner} to retrieve knowledge from the labeled and unlabeled {\ukb}s by using unconfident entities and given inputs as queries. It encodes many pieces of knowledge efficiently with sparse attention.
In experiments, {\saner} outperformed DistilBERT and BERT baselines pre-trained on the {\conll} and {\nerbert} datasets by 1.22 to 2.35 points. We found that the entity-level retrieval, the focus on the unconfident entities, the labeled and unlabeled {\ukb}s, and the large $m$ that is enabled by the sparse attention all contribute to {\saner}'s performance.

We believe that {\saner} can help application providers to develop {\ner} services in their target domain with domain-specific entity types that they have defined, even if they do not have an annotated dataset sufficiently.

\section*{Limitations}
{\saner} would be of benefit to low-resource domains and languages. 
However, for languages that have no word segmentation, such as Chinese, the method of constructing {\ukb} based on n-grams and capitalization may not be suitable.
For such languages, we can use a traditional word segmenter and POS tagger to extract entity-like n-grams.
Although we did not conduct any such data preprocessing in our experiments, it may also be useful for English.

\section*{Acknowledgement}
This work (second author) was partially supported by JSPS KAKENHI Grant Number JP21H03494. We thank all reviewers for their hard work.

\bibliography{theme}

\begin{thebibliography}{59}
\expandafter\ifx\csname natexlab\endcsname\relax\def\natexlab#1{#1}\fi

\bibitem[{Banerjee et~al.(2019)Banerjee, Pal, Devarakonda, and
  Baral}]{knowledgener}
Pratyay Banerjee, Kuntal~Kumar Pal, Murthy Devarakonda, and Chitta Baral. 2019.
\newblock \href {https://doi.org/https://doi.org/10.48550/arXiv.1911.03869}
  {Knowledge guided named entity recognition for biomedical text}.
\newblock \emph{arXiv preprint arXiv:1911.03869}.

\bibitem[{Bird et~al.(2009)Bird, Klein, and Loper}]{nltk}
Steven Bird, Ewan Klein, and Edward Loper. 2009.
\newblock \emph{Natural Language Processing with {P}ython}.
\newblock O'Reilly Media, Inc.

\bibitem[{Borgeaud et~al.(2021)Borgeaud, Mensch, Hoffmann, Cai, Rutherford,
  Millican, Driessche, Lespiau, Damoc, Clark et~al.}]{retro}
Sebastian Borgeaud, Arthur Mensch, Jordan Hoffmann, Trevor Cai, Eliza
  Rutherford, Katie Millican, George van~den Driessche, Jean-Baptiste Lespiau,
  Bogdan Damoc, Aidan Clark, et~al. 2021.
\newblock \href {https://doi.org/https://doi.org/10.48550/arXiv.2112.04426}
  {Improving language models by retrieving from trillions of tokens}.
\newblock \emph{arXiv preprint arXiv:2112.04426}.

\bibitem[{Brown et~al.(2020)Brown, Mann, Ryder, Subbiah, Kaplan, Dhariwal,
  Neelakantan, Shyam, Sastry, Askell et~al.}]{gpt-3}
Tom~B Brown, Benjamin Mann, Nick Ryder, Melanie Subbiah, Jared Kaplan, Prafulla
  Dhariwal, Arvind Neelakantan, Pranav Shyam, Girish Sastry, Amanda Askell,
  et~al. 2020.
\newblock Language models are few-shot learners.
\newblock \emph{arXiv preprint arXiv:2005.14165}.

\bibitem[{Cao et~al.(2021)Cao, Lin, Han, Sun, Yan, Liao, Xue, and Xu}]{lm_kb2}
Boxi Cao, Hongyu Lin, Xianpei Han, Le~Sun, Lingyong Yan, Meng Liao, Tong Xue,
  and Jin Xu. 2021.
\newblock \href {https://doi.org/10.18653/v1/2021.acl-long.146} {Knowledgeable
  or educated guess? revisiting language models as knowledge bases}.
\newblock In \emph{ACL-IJCNLP}, pages 1860--1874.

\bibitem[{Cao et~al.(2019)Cao, Hu, Chua, Liu, and Ji}]{ner_weak}
Yixin Cao, Zikun Hu, Tat-seng Chua, Zhiyuan Liu, and Heng Ji. 2019.
\newblock \href {https://doi.org/10.18653/v1/D19-1025} {Low-resource name
  tagging learned with weakly labeled data}.
\newblock In \emph{EMNLP-IJCNLP}, pages 261--270.

\bibitem[{Chen et~al.(2022)Chen, Li, Deng, Tan, Xu, Huang, Si, Chen, and
  Zhang}]{ner_gen4}
Xiang Chen, Lei Li, Shumin Deng, Chuanqi Tan, Changliang Xu, Fei Huang, Luo Si,
  Huajun Chen, and Ningyu Zhang. 2022.
\newblock \href {https://aclanthology.org/2022.coling-1.209} {{L}ight{NER}: A
  lightweight tuning paradigm for low-resource {NER} via pluggable prompting}.
\newblock In \emph{COLING}, pages 2374--2387.

\bibitem[{Chiu and Nichols(2016)}]{ner_data3}
Jason~P.C. Chiu and Eric Nichols. 2016.
\newblock \href {https://doi.org/10.1162/tacl_a_00104} {Named entity
  recognition with bidirectional {LSTM}-{CNN}s}.
\newblock \emph{TACL}, 4:357--370.

\bibitem[{Cohen and Sarawagi(2004)}]{gazetteer3}
William~W Cohen and Sunita Sarawagi. 2004.
\newblock \href {https://doi.org/https://doi.org/10.1145/1014052.1014065}
  {Exploiting dictionaries in named entity extraction: combining semi-{Markov}
  extraction processes and data integration methods}.
\newblock In \emph{KDD}, pages 89--98.

\bibitem[{de~Jong et~al.(2022)de~Jong, Zemlyanskiy, FitzGerald, Sha, and
  Cohen}]{tome}
Michiel de~Jong, Yury Zemlyanskiy, Nicholas FitzGerald, Fei Sha, and William
  Cohen. 2022.
\newblock \href {https://arxiv.org/abs/2110.06176} {Mention memory:
  incorporating textual knowledge into transformers through entity mention
  attention}.
\newblock In \emph{ICLR}.

\bibitem[{Devlin et~al.(2019)Devlin, Chang, Lee, and Toutanova}]{bert}
Jacob Devlin, Ming-Wei Chang, Kenton Lee, and Kristina Toutanova. 2019.
\newblock \href {https://doi.org/10.18653/v1/N19-1423} {{BERT}: Pre-training of
  deep bidirectional transformers for language understanding}.
\newblock In \emph{NAACL}, pages 4171--4186.

\bibitem[{Dhingra et~al.(2022)Dhingra, Cole, Eisenschlos, Gillick, Eisenstein,
  and Cohen}]{lm_kb3}
Bhuwan Dhingra, Jeremy~R. Cole, Julian~Martin Eisenschlos, Daniel Gillick,
  Jacob Eisenstein, and William~W. Cohen. 2022.
\newblock \href {https://doi.org/10.1162/tacl_a_00459} {Time-aware language
  models as temporal knowledge bases}.
\newblock \emph{TACL}, 10:257--273.

\bibitem[{Dodge et~al.(2021)Dodge, Sap, Marasovi{\'c}, Agnew, Ilharco,
  Groeneveld, Mitchell, and Gardner}]{case_study_c4}
Jesse Dodge, Maarten Sap, Ana Marasovi{\'c}, William Agnew, Gabriel Ilharco,
  Dirk Groeneveld, Margaret Mitchell, and Matt Gardner. 2021.
\newblock \href {https://doi.org/10.18653/v1/2021.emnlp-main.98} {Documenting
  large webtext corpora: A case study on the colossal clean crawled corpus}.
\newblock In \emph{EMNLP}, pages 1286--1305.

\bibitem[{Do{\u{g}}an et~al.(2014)Do{\u{g}}an, Leaman, and Lu}]{ner_disease}
Rezarta~Islamaj Do{\u{g}}an, Robert Leaman, and Zhiyong Lu. 2014.
\newblock \href {https://doi.org/10.1016/j.jbi.2013.12.006} {{NCBI} disease
  corpus: a resource for disease name recognition and concept normalization}.
\newblock \emph{Journal of biomedical informatics}, 47:1--10.

\bibitem[{Finkel et~al.(2005)Finkel, Grenager, and Manning}]{ner_global2}
Jenny~Rose Finkel, Trond Grenager, and Christopher Manning. 2005.
\newblock \href {https://doi.org/10.3115/1219840.1219885} {Incorporating
  non-local information into information extraction systems by {G}ibbs
  sampling}.
\newblock In \emph{ACL}, pages 363--370.

\bibitem[{Florian et~al.(2003)Florian, Ittycheriah, Jing, and
  Zhang}]{gazetteer4}
Radu Florian, Abe Ittycheriah, Hongyan Jing, and Tong Zhang. 2003.
\newblock \href {https://aclanthology.org/W03-0425} {Named entity recognition
  through classifier combination}.
\newblock In \emph{CoNLL@HLT-NAACL}, pages 168--171.

\bibitem[{Gururangan et~al.(2020)Gururangan, Marasovi{\'c}, Swayamdipta, Lo,
  Beltagy, Downey, and Smith}]{tapt}
Suchin Gururangan, Ana Marasovi{\'c}, Swabha Swayamdipta, Kyle Lo, Iz~Beltagy,
  Doug Downey, and Noah~A. Smith. 2020.
\newblock \href {https://doi.org/10.18653/v1/2020.acl-main.740} {Don{'}t stop
  pretraining: Adapt language models to domains and tasks}.
\newblock In \emph{ACL}, pages 8342--8360.

\bibitem[{Guu et~al.(2020)Guu, Lee, Tung, Pasupat, and Chang}]{realm}
Kelvin Guu, Kenton Lee, Zora Tung, Panupong Pasupat, and Mingwei Chang. 2020.
\newblock \href {https://proceedings.mlr.press/v119/guu20a.html} {Retrieval
  augmented language model pre-training}.
\newblock In \emph{ICML}, pages 3929--3938.

\bibitem[{Huang et~al.(2021)Huang, Tang, Zhong, Lu, Shou, Gong, Jiang, and
  Duan}]{bert_sentemb}
Junjie Huang, Duyu Tang, Wanjun Zhong, Shuai Lu, Linjun Shou, Ming Gong, Daxin
  Jiang, and Nan Duan. 2021.
\newblock \href {https://doi.org/10.18653/v1/2021.findings-emnlp.23}
  {{W}hitening{BERT}: An easy unsupervised sentence embedding approach}.
\newblock In \emph{Findings of EMNLP}, pages 238--244.

\bibitem[{Izacard and Grave(2021)}]{fid}
Gautier Izacard and Edouard Grave. 2021.
\newblock \href {https://doi.org/10.18653/v1/2021.eacl-main.74} {Leveraging
  passage retrieval with generative models for open domain question answering}.
\newblock In \emph{EACL}, pages 874--880.

\bibitem[{Johnson et~al.(2021)Johnson, Douze, and Jégou}]{faiss}
Jeff Johnson, Matthijs Douze, and Hervé Jégou. 2021.
\newblock \href {https://doi.org/10.1109/TBDATA.2019.2921572} {Billion-scale
  similarity search with {GPU}s}.
\newblock \emph{IEEE Transactions on Big Data}, 7(3):535--547.

\bibitem[{Kim et~al.(2003)Kim, Ohta, Tateisi, and Tsujii}]{ner_gene}
J-D Kim, Tomoko Ohta, Yuka Tateisi, and Jun’ichi Tsujii. 2003.
\newblock \href {https://doi.org/10.1093/bioinformatics/btg1023} {Genia
  corpus—a semantically annotated corpus for bio-textmining}.
\newblock \emph{Bioinformatics}, 19(suppl\_1):i180--i182.

\bibitem[{Kingma and Ba(2015)}]{adam}
Diederik~P. Kingma and Jimmy Ba. 2015.
\newblock \href {http://arxiv.org/abs/1412.6980} {Adam: A method for stochastic
  optimization}.
\newblock In \emph{ICLR (Poster)}.

\bibitem[{Krishnan and Manning(2006)}]{ner_global3}
Vijay Krishnan and Christopher~D. Manning. 2006.
\newblock \href {https://doi.org/10.3115/1220175.1220316} {An effective
  two-stage model for exploiting non-local dependencies in named entity
  recognition}.
\newblock In \emph{COLING}, pages 1121--1128.

\bibitem[{Lewis(1997)}]{reuter21578}
David~D. Lewis. 1997.
\newblock Reuters-21578 text categorization test collection, distribution 1.0.

\bibitem[{Lewis et~al.(2020)Lewis, Perez, Piktus, Petroni, Karpukhin, Goyal,
  K\"{u}ttler, Lewis, Yih, Rockt\"{a}schel, Riedel, and Kiela}]{rag}
Patrick Lewis, Ethan Perez, Aleksandra Piktus, Fabio Petroni, Vladimir
  Karpukhin, Naman Goyal, Heinrich K\"{u}ttler, Mike Lewis, Wen-tau Yih, Tim
  Rockt\"{a}schel, Sebastian Riedel, and Douwe Kiela. 2020.
\newblock \href
  {https://proceedings.neurips.cc/paper/2020/file/6b493230205f780e1bc26945df7481e5-Paper.pdf}
  {Retrieval-augmented generation for knowledge-intensive {NLP} tasks}.
\newblock In \emph{NeurIPS}, pages 9459--9474.

\bibitem[{Li et~al.(2020)Li, Feng, Meng, Han, Wu, and
  Li}]{li-etal-2020-unified}
Xiaoya Li, Jingrong Feng, Yuxian Meng, Qinghong Han, Fei Wu, and Jiwei Li.
  2020.
\newblock \href {https://doi.org/10.18653/v1/2020.acl-main.519} {A unified
  {MRC} framework for named entity recognition}.
\newblock In \emph{ACL}, pages 5849--5859.

\bibitem[{Liang et~al.(2020)Liang, Yu, Jiang, Er, Wang, Zhao, and Zhang}]{bond}
Chen Liang, Yue Yu, Haoming Jiang, Siawpeng Er, Ruijia Wang, Tuo Zhao, and Chao
  Zhang. 2020.
\newblock \href {https://doi.org/10.1145/3394486.3403149} {Bond: Bert-assisted
  open-domain named entity recognition with distant supervision}.
\newblock In \emph{KDD}, page 1054–1064.

\bibitem[{Lin and Bilmes(2011)}]{summarization_submodular}
Hui Lin and Jeff Bilmes. 2011.
\newblock \href {https://aclanthology.org/P11-1052} {A class of submodular
  functions for document summarization}.
\newblock In \emph{ACL}, pages 510--520.

\bibitem[{Liu et~al.(2019)Liu, Yao, and Lin}]{gazetteer_nn1}
Tianyu Liu, Jin-Ge Yao, and Chin-Yew Lin. 2019.
\newblock \href {https://doi.org/10.18653/v1/P19-1524} {Towards improving
  neural named entity recognition with gazetteers}.
\newblock In \emph{ACL}, pages 5301--5307.

\bibitem[{Liu et~al.(2021{\natexlab{a}})Liu, Jiang, Hu, Shi, and
  Fung}]{ner_bert}
Zihan Liu, Feijun Jiang, Yuxiang Hu, Chen Shi, and Pascale Fung.
  2021{\natexlab{a}}.
\newblock \href {https://doi.org/https://doi.org/10.48550/arXiv.2112.00405}
  {{NER-BERT}: A pre-trained model for low-resource entity tagging}.
\newblock \emph{arXiv preprint arXiv:2112.00405}.

\bibitem[{Liu et~al.(2021{\natexlab{b}})Liu, Xu, Yu, Dai, Ji, Cahyawijaya,
  Madotto, and Fung}]{crossner}
Zihan Liu, Yan Xu, Tiezheng Yu, Wenliang Dai, Ziwei Ji, Samuel Cahyawijaya,
  Andrea Madotto, and Pascale Fung. 2021{\natexlab{b}}.
\newblock \href {https://ojs.aaai.org/index.php/AAAI/article/view/17587}
  {{CrossNER}: Evaluating cross-domain named entity recognition}.
\newblock In \emph{AAAI}, pages 13452--13460.

\bibitem[{Luo et~al.(2015)Luo, Huang, Lin, and Nie}]{gazetteers2}
Gang Luo, Xiaojiang Huang, Chin-Yew Lin, and Zaiqing Nie. 2015.
\newblock \href {https://doi.org/10.18653/v1/D15-1104} {Joint entity
  recognition and disambiguation}.
\newblock In \emph{EMNLP}, pages 879--888.

\bibitem[{Luoma and Pyysalo(2020)}]{ner_corpus1}
Jouni Luoma and Sampo Pyysalo. 2020.
\newblock \href {https://doi.org/10.18653/v1/2020.coling-main.78} {Exploring
  cross-sentence contexts for named entity recognition with {BERT}}.
\newblock In \emph{COLING}, pages 904--914.

\bibitem[{Ma and Hovy(2016)}]{ner_data1}
Xuezhe Ma and Eduard Hovy. 2016.
\newblock \href {https://doi.org/10.18653/v1/P16-1101} {End-to-end sequence
  labeling via bi-directional {LSTM-CNNs-CRF}}.
\newblock In \emph{ACL}, pages 1064--1074.

\bibitem[{Mendes et~al.(2012)Mendes, Jakob, and Bizer}]{dbpedia}
Pablo Mendes, Max Jakob, and Christian Bizer. 2012.
\newblock \href
  {http://www.lrec-conf.org/proceedings/lrec2012/pdf/570_Paper.pdf} {{DB}pedia:
  A multilingual cross-domain knowledge base}.
\newblock In \emph{LREC}, pages 1813--1817.

\bibitem[{Mengge et~al.(2020)Mengge, Yu, Zhang, Liu, Zhang, and
  Wang}]{gazetteer_nn2}
Xue Mengge, Bowen Yu, Zhenyu Zhang, Tingwen Liu, Yue Zhang, and Bin Wang. 2020.
\newblock \href {https://doi.org/10.18653/v1/2020.emnlp-main.514}
  {Coarse-to-fine pre-training for named entity recognition}.
\newblock In \emph{EMNLP}, pages 6345--6354.

\bibitem[{Mintz et~al.(2009)Mintz, Bills, Snow, and Jurafsky}]{distant}
Mike Mintz, Steven Bills, Rion Snow, and Daniel Jurafsky. 2009.
\newblock \href {https://aclanthology.org/P09-1113} {Distant supervision for
  relation extraction without labeled data}.
\newblock In \emph{ACL-IJCNLP}, pages 1003--1011.

\bibitem[{Paolini et~al.(2021)Paolini, Athiwaratkun, Krone, Ma, Achille,
  ANUBHAI, dos Santos, Xiang, and Soatto}]{ner_gen1}
Giovanni Paolini, Ben Athiwaratkun, Jason Krone, Jie Ma, Alessandro Achille,
  RISHITA ANUBHAI, Cicero~Nogueira dos Santos, Bing Xiang, and Stefano Soatto.
  2021.
\newblock \href {https://openreview.net/forum?id=US-TP-xnXI} {Structured
  prediction as translation between augmented natural languages}.
\newblock In \emph{ICLR}.

\bibitem[{Paszke et~al.(2017)Paszke, Gross, Chintala, Chanan, Yang, DeVito,
  Lin, Desmaison, Antiga, and Lerer}]{pytorch}
Adam Paszke, Sam Gross, Soumith Chintala, Gregory Chanan, Edward Yang, Zachary
  DeVito, Zeming Lin, Alban Desmaison, Luca Antiga, and Adam Lerer. 2017.
\newblock \href {https://openreview.net/forum?id=BJJsrmfCZ} {Automatic
  differentiation in {PyTorch}}.
\newblock In \emph{Autodiff@NIPS}.

\bibitem[{Petroni et~al.(2019)Petroni, Rockt{\"a}schel, Riedel, Lewis, Bakhtin,
  Wu, and Miller}]{lm_kb1}
Fabio Petroni, Tim Rockt{\"a}schel, Sebastian Riedel, Patrick Lewis, Anton
  Bakhtin, Yuxiang Wu, and Alexander Miller. 2019.
\newblock \href {https://doi.org/10.18653/v1/D19-1250} {Language models as
  knowledge bases?}
\newblock In \emph{EMNLP-IJCNLP}, pages 2463--2473.

\bibitem[{Plank et~al.(2014)Plank, Hovy, McDonald, and
  S{\o}gaard}]{twitter_link}
Barbara Plank, Dirk Hovy, Ryan McDonald, and Anders S{\o}gaard. 2014.
\newblock \href {https://aclanthology.org/C14-1168} {Adapting taggers to
  {T}witter with not-so-distant supervision}.
\newblock In \emph{COLING}, pages 1783--1792.

\bibitem[{Raffel et~al.(2020)Raffel, Shazeer, Roberts, Lee, Narang, Matena,
  Zhou, Li, and Liu}]{t5}
Colin Raffel, Noam Shazeer, Adam Roberts, Katherine Lee, Sharan Narang, Michael
  Matena, Yanqi Zhou, Wei Li, and Peter~J Liu. 2020.
\newblock Exploring the limits of transfer learning with a unified text-to-text
  transformer.
\newblock \emph{JMLR}, 21:1--67.

\bibitem[{Salinas~Alvarado et~al.(2015)Salinas~Alvarado, Verspoor, and
  Baldwin}]{finance_ner}
Julio~Cesar Salinas~Alvarado, Karin Verspoor, and Timothy Baldwin. 2015.
\newblock \href {https://aclanthology.org/U15-1010} {Domain adaption of named
  entity recognition to support credit risk assessment}.
\newblock In \emph{ALTA}, pages 84--90.

\bibitem[{Sanh et~al.(2019)Sanh, Debut, Chaumond, and Wolf}]{distilbert}
Victor Sanh, Lysandre Debut, Julien Chaumond, and Thomas Wolf. 2019.
\newblock Distil{BERT}, a distilled version of {BERT}: smaller, faster, cheaper
  and lighter.
\newblock In \emph{EMC2@NeurIPS}.

\bibitem[{Seyler et~al.(2018)Seyler, Dembelova, Del~Corro, Hoffart, and
  Weikum}]{gazetteer_nn3}
Dominic Seyler, Tatiana Dembelova, Luciano Del~Corro, Johannes Hoffart, and
  Gerhard Weikum. 2018.
\newblock \href {https://doi.org/10.18653/v1/P18-2039} {A study of the
  importance of external knowledge in the named entity recognition task}.
\newblock In \emph{ACL}, pages 241--246.

\bibitem[{Shinzato et~al.(2022)Shinzato, Yoshinaga, Xia, and
  Chen}]{retreive_train2}
Keiji Shinzato, Naoki Yoshinaga, Yandi Xia, and Wei-Te Chen. 2022.
\newblock \href {https://doi.org/10.18653/v1/2022.acl-short.25} {Simple and
  effective knowledge-driven query expansion for {QA}-based product attribute
  extraction}.
\newblock In \emph{ACL}, pages 227--234.

\bibitem[{Singh et~al.(2021)Singh, Reddy, Hamilton, Dyer, and Yogatama}]{emdr}
Devendra Singh, Siva Reddy, Will Hamilton, Chris Dyer, and Dani Yogatama. 2021.
\newblock \href
  {https://proceedings.neurips.cc/paper/2021/file/da3fde159d754a2555eaa198d2d105b2-Paper.pdf}
  {End-to-end training of multi-document reader and retriever for open-domain
  question answering}.
\newblock In \emph{NeurIPS}, pages 25968--25981.

\bibitem[{Sutton and McCallum(2004)}]{ner_global1}
Charles Sutton and Andrew McCallum. 2004.
\newblock Collective segmentation and labeling of distant entities in
  information extraction.
\newblock In \emph{SRL@ICML}.

\bibitem[{Tjong Kim~Sang and De~Meulder(2003)}]{conll03}
Erik~F. Tjong Kim~Sang and Fien De~Meulder. 2003.
\newblock \href {https://aclanthology.org/W03-0419} {Introduction to the
  {CoNLL}-2003 shared task: Language-independent named entity recognition}.
\newblock In \emph{CoNLL}, pages 142--147.

\bibitem[{Trieu et~al.(2022)Trieu, Miwa, and Ananiadou}]{distant_bio}
Hai-Long Trieu, Makoto Miwa, and Sophia Ananiadou. 2022.
\newblock \href {https://doi.org/10.18653/v1/2022.bionlp-1.17} {Named entity
  recognition for cancer immunology research using distant supervision}.
\newblock In \emph{BioNLP@ACL}, pages 171--177.

\bibitem[{Virtanen et~al.(2019)Virtanen, Kanerva, Ilo, Luoma, Luotolahti,
  Salakoski, Ginter, and Pyysalo}]{ner_corpus2}
Antti Virtanen, Jenna Kanerva, Rami Ilo, Jouni Luoma, Juhani Luotolahti, Tapio
  Salakoski, Filip Ginter, and Sampo Pyysalo. 2019.
\newblock \href {https://doi.org/https://doi.org/10.48550/arXiv.1912.07076}
  {Multilingual is not enough: {BERT} for {Finnish}}.
\newblock \emph{arXiv preprint arXiv:1912.07076}.

\bibitem[{Wang et~al.(2022)Wang, Xu, Fang, Liu, Sun, Xu, Zhu, and
  Zeng}]{retreive_train1}
Shuohang Wang, Yichong Xu, Yuwei Fang, Yang Liu, Siqi Sun, Ruochen Xu,
  Chenguang Zhu, and Michael Zeng. 2022.
\newblock \href {https://doi.org/10.18653/v1/2022.acl-long.226} {Training data
  is more valuable than you think: A simple and effective method by retrieving
  from training data}.
\newblock In \emph{ACL}, pages 3170--3179.

\bibitem[{Wang et~al.(2021)Wang, Hu, Song, Garg, Xiao, and Han}]{distant_chem}
Xuan Wang, Vivian Hu, Xiangchen Song, Shweta Garg, Jinfeng Xiao, and Jiawei
  Han. 2021.
\newblock \href {https://doi.org/10.18653/v1/2021.emnlp-main.424} {{C}hem{NER}:
  Fine-grained chemistry named entity recognition with ontology-guided distant
  supervision}.
\newblock In \emph{EMNLP}, pages 5227--5240.

\bibitem[{Wang et~al.(2020)Wang, Song, Li, Guan, and Han}]{ner_covid19}
Xuan Wang, Xiangchen Song, Bangzheng Li, Yingjun Guan, and Jiawei Han. 2020.
\newblock \href {https://doi.org/https://doi.org/10.48550/arXiv.2003.12218}
  {Comprehensive named entity recognition on {CORD}-19 with distant or weak
  supervision}.
\newblock \emph{arXiv preprint arXiv:2003.12218}.

\bibitem[{Wolf et~al.(2020)Wolf, Debut, Sanh, Chaumond, Delangue, Moi, Cistac,
  Rault, Louf, Funtowicz, Davison, Shleifer, von Platen, Ma, Jernite, Plu, Xu,
  Scao, Gugger, Drame, Lhoest, and Rush}]{transformers}
Thomas Wolf, Lysandre Debut, Victor Sanh, Julien Chaumond, Clement Delangue,
  Anthony Moi, Pierric Cistac, Tim Rault, Rémi Louf, Morgan Funtowicz, Joe
  Davison, Sam Shleifer, Patrick von Platen, Clara Ma, Yacine Jernite, Julien
  Plu, Canwen Xu, Teven~Le Scao, Sylvain Gugger, Mariama Drame, Quentin Lhoest,
  and Alexander~M. Rush. 2020.
\newblock \href {https://www.aclweb.org/anthology/2020.emnlp-demos.6}
  {Transformers: State-of-the-art natural language processing}.
\newblock In \emph{ACL: System Demonstrations}, pages 38--45.

\bibitem[{Yadav and Bethard(2018)}]{ner_data4}
Vikas Yadav and Steven Bethard. 2018.
\newblock \href {https://aclanthology.org/C18-1182} {A survey on recent
  advances in named entity recognition from deep learning models}.
\newblock In \emph{COLING}, pages 2145--2158.

\bibitem[{Yan et~al.(2021)Yan, Gui, Dai, Guo, Zhang, and Qiu}]{ner_gen2}
Hang Yan, Tao Gui, Junqi Dai, Qipeng Guo, Zheng Zhang, and Xipeng Qiu. 2021.
\newblock \href {https://doi.org/10.18653/v1/2021.acl-long.451} {A unified
  generative framework for various {NER} subtasks}.
\newblock In \emph{ACL-IJCNLP}, pages 5808--5822.

\bibitem[{Zhang et~al.(2022)Zhang, Shen, Tan, Wu, and Lu}]{ner_gen3}
Shuai Zhang, Yongliang Shen, Zeqi Tan, Yiquan Wu, and Weiming Lu. 2022.
\newblock \href {https://doi.org/10.18653/v1/2022.acl-long.59} {De-bias for
  generative extraction in unified {NER} task}.
\newblock In \emph{ACL}, pages 808--818.

\end{thebibliography}
\bibliographystyle{acl_natbib}

\appendix

\section{Experimental Setup}
\label{append:hyper}
Table~\ref{tab:data} shows the data statistics. Because the finance dataset provides no development data, we split the front half of the 306 test 
examples into our development split and the back half into our test split.

We collected the raw text in the finance domain from Wikipedia articles.
We used the dump data of Wikipedia Circus Search.\footnote{\url{https://dumps.wikimedia.org/other/cirrussearch/}}
The articles in the data are automatically annotated with topic information, and we extracted the articles whose topics include ``Business and Economics'' and used them as the articles in the finance domain.

The text encoder and tokenizer were the pre-trained \textsc{bert}-base-cased model (110M parameters). 
The pre-training took 17 hours on eight NVIDIA Quadro RTX 8000 (48GB) GPUs.
The training of the largest CoNLL dataset took 6 hours on one GPU\@.
The hyperparameter settings are listed in Table~\ref{tab:hyper1}. 
We set the early stop epoch to five only in CoNLL03 for computational efficiency.
We used the Adam optimizer~\cite{adam}, PyTorch (ver.~1.10.1)\footnote{\url{https://pytorch.org/}}~\cite{pytorch}, and transformers (ver.~4.15.0)\footnote{\url{https://github.com/huggingface/transformers}}~\cite{transformers}.
Stop words were implemented with NLTK (ver.~3.7)\footnote{\url{https://www.nltk.org/}}~\cite{nltk}.
We used faiss (ver.~1.7.2)\footnote{\url{https://github.com/facebookresearch/faiss}}~\cite{faiss} for the nearest-neighbor search in the knowledge retrieval.
We set $L =64$ for all of the data preprocessing, with a sliding window size of 16.
For entities in the sliding window, we used the max operation to select from the two predictions.

We pre-trained the {\nerbert} model under the same hyperparameter settings as above, without knowledge retrieval (that is, $m=0$). This pre-training was the different from the original  {\nerbert} in terms of the sequence segmentation, initialization, and data collection results, in addition to the hyperparameters.

\begin{table}[t!]
\centering
    \footnotesize
		\begin{tabular}{lccccc} \toprule
        		& \# Train & \# Dev & \# Test & \# Types & UKB 
        		\\ \midrule
                AI. & 100 & 350 & 431 & 14 & 15 \\
                Mus. & 100 & 380 & 456 & 13 & 467 \\
                Lit. & 100 & 400 & 416 & 12 & 436 \\
                Sci. & 200 & 450 & 543 & 17 & 191 \\
                Pol. & 200 & 541 & 651 & 9 & 354 \\
                Fin. & 1169 & 103 & 103 & 4 & 850 \\
                {\conll} & 14987 & 3466 & 3684 & 4 & 7.5 \\
				 \bottomrule
				\end{tabular}
	\caption{Data Statistics. \textsc{UKB} indicates the size of \textsc{UKB} (MB).}
	\label{tab:data}
\end{table}

\begin{table}[t!]
\centering
    \footnotesize
		\begin{tabular}{ccc}\toprule
                & Pre-Training & Fine-Tuning \\ \midrule
				 Batch size & 1024 & 16 \\
				 \# Epochs & 1 & 300 \\
				 \# Steps & 10000 & --- \\
                 \# Early stop & --- & 5/8 \\
				 $m$ & 2 & 10 \\
				 $n$ & 3 & 3 \\
				 $\lambda_\mathrm{conf}$ & --- & 0.9 \\
				 $\lambda_{1}$ & --- & 0.1 \\
				 Learning rate & 5e-5 &  5e-5 \\
				 \bottomrule
				\end{tabular}
	\caption{Hyperparameters.}
	\label{tab:hyper1}
\end{table}

\section{Our implementation of {\nerbert}}
\label{append:data}

\paragraph{Data Collection}
We used the Wikipedia dump on 27, Jan., 2022 and the DBPedia Ontlogy dump on 1.\ Dec.\  2021.\footnote{We used en-specific data, which means that the types are annotated without transitive augmentation. \url{https://databus.dbpedia.org/dbpedia/mappings/instance-types/}}
Then, we split the corpus into fixed-length token sequences and removed the sequences without entities that were not labeled as ``ENTITY.'' 

We reduced the proportion of ``ENTITY'' labels by using filtering rules and down sampling.
We randomly filtered the sentences to reduce these labels. If all entities in a sentence were the top-20 frequent labels, the sentences were randomly removed from the dataset: 30\% if the number of ``ENTITY'' entities was three, 50\% if the number was four, and 70\% if the number was more than four.
In the pre-training, we used weighted sampling.
The sampling weight of the sentence was $\min_{0 \leq i \leq l} |E_{c_i} |^{-0.3}$, where $E_c$ is the number of entities of type $c$ in the dataset, and $c_i$ is the type of the $i$-th token. 
As a result, the final dataset had 33M examples, 939M tokens, and 404 types.\footnote{\citet{ner_bert} reported their data has 16.3M examples, 457.6M tokens, and 315 types. However, they had not published their data or the URLs of the dump data before our experiments.}
With the exception of the loss function, initialization, and the use of the retrieval-augmented model, we followed the procedure of the {\nerbert} pre-training algorithm.

\paragraph{Loss Function}
In addition to the cross-entropy loss used in the original {\nerbert}, we incorporated a multi-task loss to efficiently learn the {\ner} ability by ignoring the very frequent ``ENTITY'' type in the entity typing.
For the entity extraction, we performed three-class classification tasks.
We summed the output probabilities of the final linear layer after the softmax activation to obtain the probabilities of ``B-[type]'', ``I-[type]'', and ``O.'' 
In the entity typing, we masked the output logits of the final linear layer corresponding to the ``ENTITY'' label. Then, we performed the $2C_{\mathrm{pre}} -1$ classification task.
The total loss was the sum of the two cross-entropy losses. 

\paragraph{Initialization}
We had to initialize the weight of the final linear layer and the token-type embeddings because of the mismatch of the set of the labels between the downstream and pre-training tasks.
Instead of a random initialization from $\mathcal{N}(0, \sigma_0)$, where $\sigma_0 \in \mathbb{R}$ is a fixed standard deviation, we used the learned distribution $\mathcal{N}(\bm{\mu}, \bm{\sigma})$, where $\bm{\mu}, \bm{\sigma} \in \mathbb{R}^d$ is the bias and the standard deviation of the weight of the final linear layer and the token-type embeddings in the pre-trained model.

\section{Summarization-Based Filtering}
\label{append:summarize}
To assign n-gram keys to each piece of knowledge, we removed those n-grams that had any stop words or had no capital letter, so as to collect entity-like n-grams.
In addition, we used filtering methods based on the string matching and the extractive summarization.
The summarization-based filtering enabled us to limit the number of n-grams in each piece of knowledge.

We formulated the extraction of a fixed number of representative n-grams from a sequence as an extractive summarization task, as follows.
Here, let $\bm{h}_i$ be an n-gram embedding whose start position is $i$, regardless of whether the n-gram is filtered out or not.
$\bm{S} \in \mathbb{R}^{L \times L}$ is the cosine similarity matrix of $\bm{h}_i~(0\leq i <L)$. 
We denote the token spans as $\{I_s\}$; each span is a maximal token span that does not include stop words but includes a capital letter. We should extract n-grams from different spans to increase the diversity of n-grams. $\mathcal{I}_s$ is the set of such spans.

We defined the optimization problem as follows:
$Z \subseteq \{0, 1, \cdots, L-1 \}$ denotes the set of n-grams.
We used a sub-modular function as the objective to be maximized, under the constraint $|Z| \leq N_{max}$~\cite{summarization_submodular}. The objective function is
\[
\mathcal{L}_{\mathrm{cov}}(Z) =\sum_{0\leq i < L} \min \left( \sum_{j \in Z} s_{ij}, \alpha \sum_{0\leq k < L} s_{ik} \right),
\]
\[
\mathcal{L}_{\mathrm{div}}(Z) =\sum_{I_s \in \mathcal{I}_s} \sqrt{\sum_{j \in Z \cap I_s} \left( \dfrac{1}{L} \sum_{0\leq i < L} s_{ij}\right) },
\]
\[
\mathcal{L}_{\mathrm{sum}}(Z) =\mathcal{L}_{\mathrm{cov}}(Z) + \lambda_{\mathrm{div}} \mathcal{L}_{\mathrm{div}}(Z). 
\]
The hyperparameters are $\alpha =0.1, \lambda_{\mathrm{div}} =10$, and $N_{\mathrm{max}} =3$. We also required $Z$ to meet the filtering condition (that is, the inclusion of a capital letter and no stop word). $\mathcal{L}_{\mathrm{cov}}(Z)$ measures the coverage of the n-grams and $\mathcal{L}_{\mathrm{div}}(Z)$ measures the diversity of the n-grams.

Because this objective function is a sub-modular function, the greedy algorithm has a $(1-1/e)$ approximation guarantee. Therefore, we can use a lightweight computation to extract the most important n-grams.


\section{Effect of Overlapping Entities}
\label{append:overlap}
\begin{table}[t!]
\centering
	\small
		\begin{tabular}{@{}l@{\,\,\,}c@{\,\,\,}c@{}} \toprule
		        Method & Acc & $\Delta$ \\ \midrule
                \textsc{bert} on {\conll} & 70.16 {\scriptsize (0.56)} & \\ \cmidrule(l{1pt}r{0pt}){1-3}
                 \textsc{bert} on {\nerbert} (non-overlap) & 73.59 {\scriptsize (0.19)} & 3.43 \\ 
                {\saner} on {\nerbert} (non-overlap) & 75.13 {\scriptsize (0.19)} & 4.97\\  \cmidrule(l{1pt}r{0pt}){1-3}
                \textsc{bert} on {\nerbert} (overlap) & 75.90 {\scriptsize (0.19)} & 5.74 \\
                {\saner} on {\nerbert} (overlap) & 77.33 {\scriptsize (0.19)} & 7.17 \\
                \bottomrule
                \end{tabular}
	\caption{Performance on the development set. The models were pre-trained on CoNLL03, {\nerbert} without the entity overlap, and {\nerbert} with the entity overlap.}
	\label{tab:overlap}
\end{table}
To confirm that the effectiveness of {\nerbert} is not due to the overlapping entities in the pre-training and fine-tuning dataset, we conducted experiments where we removed sequences including the entities that appeared in the CrossNER dataset from the {\nerbert} corpus. Table~\ref{tab:overlap} shows the results. We confirmed that the NER ability learned from the {\nerbert} corpus itself improved performance and {\saner} outperformed {\nerbert} in both settings.

However, we also found that the performance of {\nerbert} is overestimated because of entity overlap. \citet{gpt-3} and \citet{case_study_c4} also noted that leakage of the benchmark datasets from the pre-training corpus affects the performance of GPT-3~\cite{gpt-3} and T5~\cite{t5}. The community should solve this problem in future.

\end{document}